\documentclass{article}
\usepackage{spconf,amsmath,graphicx,hyperref}
\usepackage{amssymb}  

\title{Adaptive-VoCo: Complexity-Aware Visual Token Compression for Vision-Language Models}
\name{Xiaoyang Guo, Keze Wang}
\address{Sun Yat-sen University}

\begin{document}
%
\maketitle
\begin{abstract}
In recent years, large-scale vision-language models (VLMs) have demonstrated remarkable performance on multimodal understanding and reasoning tasks. However, handling high-dimensional visual features often incurs substantial computational and memory costs. VoCo-LLaMA alleviates this issue by compressing visual patch tokens into a few VoCo tokens, reducing computational overhead while preserving strong cross-modal alignment. Nevertheless, such approaches typically adopt a fixed compression rate, limiting their ability to adapt to varying levels of visual complexity. To address this limitation, we propose Adaptive-VoCo, a framework that augments VoCo-LLaMA with a lightweight predictor for adaptive compression. This predictor dynamically selects an optimal compression rate by quantifying an image's visual complexity using statistical cues from the vision encoder, such as patch token entropy and attention map variance. Furthermore, we introduce a joint loss function that integrates rate regularization with complexity alignment. This enables the model to balance inference efficiency with representational capacity, particularly in challenging scenarios. Experimental results show that our method consistently outperforms fixed-rate baselines across multiple multimodal tasks, highlighting the potential of adaptive visual compression for creating more efficient and robust VLMs.
\end{abstract}
\section{Introduction}
\label{sec:intro}

Recent progress in large-scale vision-language models (VLMs) has greatly improved multimodal understanding, enabling applications such as image captioning, visual question answering, and cross-modal reasoning \cite{z1,z2, liu2024llavanext, Qwen2.5-VL}. This success relies on rich visual token representations that capture fine-grained image details, but high-resolution inputs can generate thousands of tokens, saturating the limited context window of large language models (LLMs) and constraining text-based reasoning. To address this bottleneck, researchers have explored visual token compression strategies that condense dense patch-level embeddings into a compact set of tokens. Early methods include Q-Former \cite{li2023blip} and average pooling \cite{z3}, while VoCo-LLaMA \cite{z4} integrates compression directly into the language model via learnable VoCo tokens and attention distillation, achieving high compression rates with minimal performance loss.

Despite these improvements, current compression approaches rely on a fixed token budget that is applied uniformly across all inputs. Such rigidity overlooks the inherent variability in visual complexity: while simple scenes may be sufficiently represented with very few tokens, complex or high-resolution images demand richer representations. Consequently, fixed-rate compression can discard crucial details in challenging cases, limiting the adaptability and overall effectiveness of existing models.

To overcome this limitation, we propose \textbf{Adaptive-VoCo}, a framework that introduces a dynamic, complexity-aware approach to visual token compression. At the core of our method is a lightweight \emph{Rate Predictor} module that estimates the intrinsic visual complexity of each input. It analyzes a rich set of signals from the vision encoder—such as the statistical distribution of patch embeddings and the dispersion of attention maps—to produce a complexity score. Based on this score, the predictor dynamically allocates an appropriate number of \texttt{<voco>} tokens from a discrete candidate set (e.g., \{1, 2, 4\}). This allows simple images to be represented with maximum efficiency while granting more representational capacity to complex scenes. To guide this allocation policy, we design a novel, complexity-aware training objective that encourages the model to be efficient by default while ensuring the token budget aligns with the image's difficulty. This objective is optimized jointly with the standard language modeling loss, enabling the model to learn both the primary task and an efficient resource allocation strategy in an end-to-end fashion.

In summary, our contributions are threefold: (1) we introduce Adaptive-VoCo, the first complexity-aware visual token compression framework that adaptively allocates token budgets per input; (2) we design a lightweight yet effective Rate Predictor that leverages visual statistics and attention signals to guide token allocation; and (3) we propose a novel loss formulation that balances computational efficiency with representational fidelity, achieving advanced performance across multiple vision-language benchmarks.

\section{Related Work}
\label{sec:related_work}
\subsection{Evolution of Fixed-Rate Compression}
The challenge of managing extensive visual token sequences in multimodal Large Language Models (LLMs) has spurred the development of various compression techniques. Early strategies often relied on external modules, such as the query-based mechanism in Q-Former  or simple average pooling \cite{li2024llama}, to map a variable number of patch tokens to a fixed-length representation. While effective at reducing computational load, these methods are often insensitive to the visual complexity of the input, leading to significant performance degradation on detailed or intricate scenes \cite{z5, li2024llama, ye2025voco, z6}. A notable advancement came with VoCo-LLaMA \cite{ye2025voco}, which, inspired by text compression techniques \cite{z8, jiang2023llmlingua, z7}, integrates compression directly into the LLM. By using learnable tokens and attention distillation, VoCo-LLaMA achieves high compression rates while effectively preserving visual information, demonstrating the potential of leveraging the model's own architecture for compression.
\subsection{Limitation of Static Token Allocation}
Despite the progress, a common limitation persists: these methods \cite{ye2025voco, zhang2025llava, z12, z10}, including VoCo-LLaMA \cite{ye2025voco}, operate with a \textit{fixed} compression rate. This static approach applies the same token budget to all images, regardless of whether a scene is simple and easily compressible or complex and requiring a richer representation. This inflexibility creates a critical performance trade-off, as a single, predetermined token count is inherently suboptimal for handling the diverse complexity of real-world visual data. Our work directly addresses this gap by introducing a dynamic, complexity-aware compression mechanism that adapts the token budget on a per-image basis.
\section{Method}
\label{sec:method}
\begin{figure*}[t]
  \centering
  \includegraphics[width=0.8\textwidth]{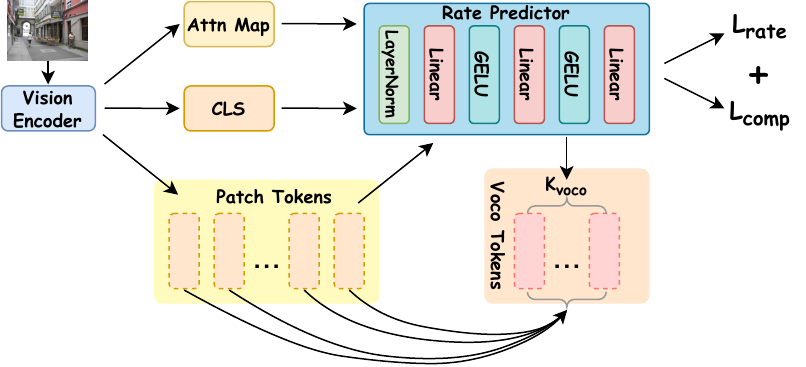}
  \caption{Overview of the proposed Adaptive-VoCo framework. The input image is processed by a vision encoder to extract the class token (CLS), patch-level embeddings, and the attention map. These features are fed into the Rate Predictor to estimate the appropriate number of compressed visual tokens $K_{\text{voco}}$. The predicted token number is then used to expand the placeholder token into $K_{\text{voco}}$ VoCo tokens. The Rate Predictor is trained with the Rate Loss ($L_{\text{rate}}$) and the Complexity Loss ($L_{\text{comp}}$) to ensure efficient and complexity-aware token allocation.}
  \label{fig:pipeline}
\end{figure*}
\textbf{Adaptive-VoCo Overview:}
Existing Vision-Language Models (VLMs) typically employ a static visual token compression strategy, mapping a high-dimensional set of patch embeddings to a representation of a fixed, predetermined size. This approach fails to account for the intrinsic variability in visual complexity across different images. To address this limitation, we introduce Adaptive-VoCo, a framework that recasts visual token compression as a dynamic resource allocation problem. Our central innovation is a lightweight, learnable \textbf{Rate Predictor}, which formulates a policy $\pi(K|\mathbf{I})$ to select an appropriate number of compressed tokens $K$ for a given image $\mathbf{I}$. This policy is conditioned on a set of statistical indicators extracted from the vision encoder, including token distribution properties and attention map dispersion. By dynamically allocating more tokens to visually complex inputs and fewer to simpler ones from a discrete candidate set $\mathcal{K} = \{k_1, k_2, \dots, k_{|\mathcal{K}|}\}$, Adaptive-VoCo optimizes the trade-off between computational efficiency and representational fidelity. The overall architecture is depicted in Figure~\ref{fig:pipeline}.

\subsection{Complexity-Aware Rate Prediction}
\label{ssec:rate_predictor}

The Rate Predictor module, denoted as $\mathcal{R}_\theta$ with parameters $\theta$, is designed to infer the optimal token count $K$ by estimating an input's visual complexity.

\noindent\textbf{Feature Extraction.}
Given an input image $\mathbf{I}$, a vision encoder $\mathcal{E}_{\text{vision}}$ produces a sequence of $N$ patch embeddings $\mathbf{P} = \{\mathbf{p}_i\}_{i=1}^N \subset \mathbb{R}^{d}$ and the final layer's self-attention map $\mathbf{A} \in \mathbb{R}^{N \times N}$. We derive a complexity feature vector $\mathbf{f}_{\text{comp}}$ by aggregating several statistical measures. First, we compute the empirical mean and variance of the patch embeddings:
\begin{align}
\scalebox{1.0}{$
    \boldsymbol{\mu}_{\mathbf{P}} = \frac{1}{N} \sum_{i=1}^N \mathbf{p}_i
$} \\
\scalebox{0.9}{$
    \boldsymbol{\sigma}_{\mathbf{P}}^2 = \frac{1}{N} \sum_{i=1}^N (\mathbf{p}_i - \boldsymbol{\mu}_{\mathbf{P}})^2
$}
\end{align}
where the square is applied element-wise. These statistics are concatenated with scalar measures of feature diversity and global structure—patch token entropy $H(\mathbf{P})$ and attention variance $\text{Var}(\mathbf{A})$—to form the final feature vector:
\begin{equation}
\mathbf{f}_{\text{comp}} = [\boldsymbol{\mu}_{\mathbf{P}}^\top, (\boldsymbol{\sigma}_{\mathbf{P}}^2)^\top, H(\mathbf{P}), \text{Var}(\mathbf{A})]^\top \in \mathbb{R}^{2d+2}.
\end{equation}
The patch token entropy $H(\mathbf{P})$ is defined using a softmax over the L2 norms of the patch embeddings, which serves as a proxy for token salience:
\begin{equation}
\small
H(\mathbf{P}) = -\sum_{i=1}^N q_i \log q_i, \quad \text{where} \quad q_i = \frac{\exp(\|\mathbf{p}_i\|_2 / \tau_e)}{\sum_{j=1}^N \exp(\|\mathbf{p}_j\|_2 / \tau_e)},
\label{eq:entropy}
\end{equation}
with $\tau_e$ as a temperature hyperparameter. 

\noindent\textbf{Probabilistic Rate Selection.}
The feature vector $\mathbf{f}_{\text{comp}}$ is processed by $\mathcal{R}_\theta$, a multi-layer perceptron (MLP), to produce logits $\mathbf{z} \in \mathbb{R}^{|\mathcal{K}|}$ over the candidate token counts. We then obtain a probability distribution $\boldsymbol{\pi} = \{\pi_j\}_{j=1}^{|\mathcal{K}|}$ via the softmax function: $\boldsymbol{\pi} = \text{softmax}(\mathbf{z})$.

To maintain differentiability during end-to-end training, we employ the Gumbel-Softmax trick to sample a token count $K$ from this distribution. A categorical sample $\mathbf{y} \in \{0,1\}^{|\mathcal{K}|}$ is drawn as:
\begin{equation}
\mathbf{y}_j = \frac{\exp((\log(\pi_j) + g_j)/\tau_g)}{\sum_{l=1}^{|\mathcal{K}|} \exp((\log(\pi_l) + g_l)/\tau_g)},
\end{equation}
where $g_j = -\log(-\log(u_j))$ with $u_j \sim \text{Uniform}(0,1)$, and $\tau_g$ is the Gumbel temperature. The selected token count is $K = \mathbf{y}^\top \mathbf{k}$, where $\mathbf{k}$ is the vector of candidate counts. At inference, we deterministically select the count with the highest probability: $K = k_{\arg\max(\mathbf{z})}$.

\subsection{Adaptive Token Allocation}
The predicted token count $K$ dynamically modulates the structure of the input sequence fed to the multimodal transformer. We designate a special placeholder token, \texttt{<voco>}, in the initial prompt. This token is then expanded into a sequence of $K$ consecutive \texttt{<voco>} tokens. This transformation can be formally represented as an operator $\mathcal{T}$ acting on the token sequence $\mathbf{S}$:
\begin{equation}
\mathcal{T}(\mathbf{S}, K) = \mathbf{S}_{\text{pre}} \oplus (\underbrace{\texttt{<voco>}, \dots, \texttt{<voco>}}_{K \text{ tokens}}) \oplus \mathbf{S}_{\text{post}},
\end{equation}
where $\oplus$ denotes concatenation, and $\mathbf{S}_{\text{pre}}$ and $\mathbf{S}_{\text{post}}$ are the token subsequences preceding and succeeding the original placeholder. This mechanism ensures that the model's representational capacity is allocated proportionally to the input's visual complexity while preserving positional consistency for subsequent transformer layers.

\subsection{Loss Formulation for Rate Predictor Training}
\label{sec:loss_design}
The training of the Rate Predictor $\mathcal{R}_\theta$ is guided by a dual-objective loss function designed to balance computational parsimony with representational accuracy.

\textbf{Rate Loss ($L_{\text{rate}}$).} To encourage efficiency, the Rate Loss penalizes the excessive allocation of tokens. It is formulated based on the expected number of tokens, normalized by the maximum possible count $k_{\max} = \max(\mathcal{K})$:
\begin{equation}
\scalebox{0.9}{$
L_{\text{rate}} = \frac{1}{B \cdot k_{\max}} \sum_{i=1}^{B} \left( \sum_{j=1}^{|\mathcal{K}|} \pi_j^{(i)} k_j \right)
$}
\end{equation}
where $B$ is the batch size and $\pi_j^{(i)}$ is the predicted probability for the $j$-th token count for the $i$-th sample. Minimizing $L_{\text{rate}}$ pushes the model towards a more compact default representation.

\textbf{Complexity Loss ($L_{\text{comp}}$).} To ensure that the allocated token count correlates with visual complexity, we introduce a Complexity Loss. We first define a target complexity score $C \in [0, 1]$ as a weighted combination of normalized information-theoretic measures:
\begin{equation}
C = \alpha \cdot \frac{\log(1 + H(\mathbf{P}))}{\log(1 + \log N)} + \beta \cdot \frac{\log(1 + \text{Var}(\mathbf{A}))}{\gamma_A},
\end{equation}
where $\alpha$ and $\beta$ are weighting hyperparameters. The entropy is normalized by its theoretical maximum $\log N$ for a discrete distribution over $N$ items. The attention variance is normalized by an empirical constant $\gamma_A$. The loss is the mean squared error between the normalized expected token count and this target score:
\begin{equation}
\scalebox{0.9}{$
L_{\text{comp}} = \frac{1}{B} \sum_{i=1}^{B} \left( \frac{1}{k_{\max}} \sum_{j=1}^{|\mathcal{K}|} \pi_j^{(i)} k_j - C^{(i)} \right)^2
$}
\end{equation}
This objective explicitly encourages $\mathcal{R}_\theta$ to learn a policy that allocates a larger token budget to inputs with higher entropy and more diffuse attention patterns, which are proxies for visual complexity.
\section{Experiments}
\label{sec:experiments}
\begin{table*}[t!]
\centering
\resizebox{\textwidth}{!}{%
\begin{tabular}{|l|c|c|c|c|c|c|c|c|c|}
\hline
\textbf{Model} & \textbf{Tokens} & \textbf{GQA} & \textbf{VQAv2} & \textbf{MMBench} & \textbf{MME} & \textbf{POPE} & \textbf{SEED} & \textbf{SQA$^I$} & \textbf{Avg. (\%)} \\
\hline
Upper Bound & 576 & 61.1 & 77.7 & 64.0 & 1487.2 & 85.0 & 57.9 & 66.5 & 100.0 \\
\hline
Q-Former \cite{li2023blip} & 1 & 51.1 & 63.4 & 51.7 & 1079.7 & 77.3 & 47.2 & 62.7 & 57.2 \\
Avg. Pool \cite{li2024llama} & 1 & 52.9 & 65.0 & 55.5 & 1210.3 & 79.1 & 50.3 & 62.2 & 64.1 \\
VoCo-LLaMA \cite{ye2025voco} & 1 & 57.4 & 71.8 & 57.9 & 1241.4 & \textbf{81.5} & 48.8 & 66.3 & 81.0 \\
\textbf{Adaptive-VoCo (ours)} & \textbf{$1\!-\!4$} & \textbf{57.6} & \textbf{72.3} & \textbf{60.7} & \textbf{1285.1} & 81.4 & \textbf{50.2} & \textbf{68.5} & \textbf{89.3} \\
\hline
Lower Bound & 1 & 37.7 & 41.2 & 22.3 & 617.3 & 53.9 & 36.9 & 60.7 & 0.0 \\
\hline
\end{tabular}
}
\caption{Results on multimodal benchmarks. 
The last column (Avg.) reports the average normalized performance retention rate across all benchmarks, where the retention rate for each benchmark is computed as
$\frac{\text{Result} - \text{Lower Bound}}{\text{Upper Bound} - \text{Lower Bound}}$..}
\label{tab:benchmark_results}
\end{table*}
\subsection{Experimental Setup} To ensure a fair comparison with VoCo-LLaMA \cite{ye2025voco}, we adopt its visual encoder, language model, training and inference procedures, and data preprocessing. We evaluate on seven standard visual understanding benchmarks: GQA \cite{hudson2019gqa}, MMBench \cite{liu2024mmbench}, MME \cite{yin2024survey}, POPE \cite{li2023evaluating}, SEED-Bench \cite{li2023seed}, SQA$^I$ (image modality) \cite{lu2022learn}, and VQAv2 \cite{goyal2017making}. Baselines include Q-Former \cite{li2023blip}, Avg. Pooling \cite{li2024llama}, and fixed-rate VoCo-LLaMA, as well as Upper and Lower Bound models representing optimal and minimal compression performance. Results for Q-Former and Avg. Pooling are taken from VoCo-LLaMA, while we reproduce VoCo-LLaMA results following the original settings. Our proposed \textbf{Adaptive-VoCo} dynamically selects the number of visual tokens based on input complexity. Performance is measured using standard task metrics and a compression retention rate, reflecting the trade-off between visual token reduction and preserved model accuracy.

\subsection{Main Results} 
Table~\ref{tab:benchmark_results} summarizes the performance of our Adaptive-VoCo model compared with representative compression baselines across seven multimodal benchmarks. External aggregation methods such as Q-Former \cite{li2023blip} and Avg. Pool \cite{li2024llama} suffer from severe performance degradation under aggressive compression, while fixed-rate VoCo-LLaMA \cite{ye2025voco} achieves a much stronger average retention of $81.0\%$ by leveraging the language model itself for token compression. 

Building on this, our proposed \textbf{Adaptive-VoCo} further improves performance, achieving the highest average retention score of \textbf{89.3\%}. Notably, Adaptive-VoCo consistently outperforms the fixed-rate counterpart across most benchmarks, including GQA (+0.2), VQAv2 (+0.5), MMBench (+2.8), MME (+43.7), SEED (+1.4), and SQA$^I$ (+2.2). Even in POPE, where the results are close, Adaptive-VoCo matches the strong baseline performance.
\begin{figure}[htb]
\begin{minipage}[b]{0.49\linewidth}
  \centering
  \centerline{\includegraphics[width=4.0cm]{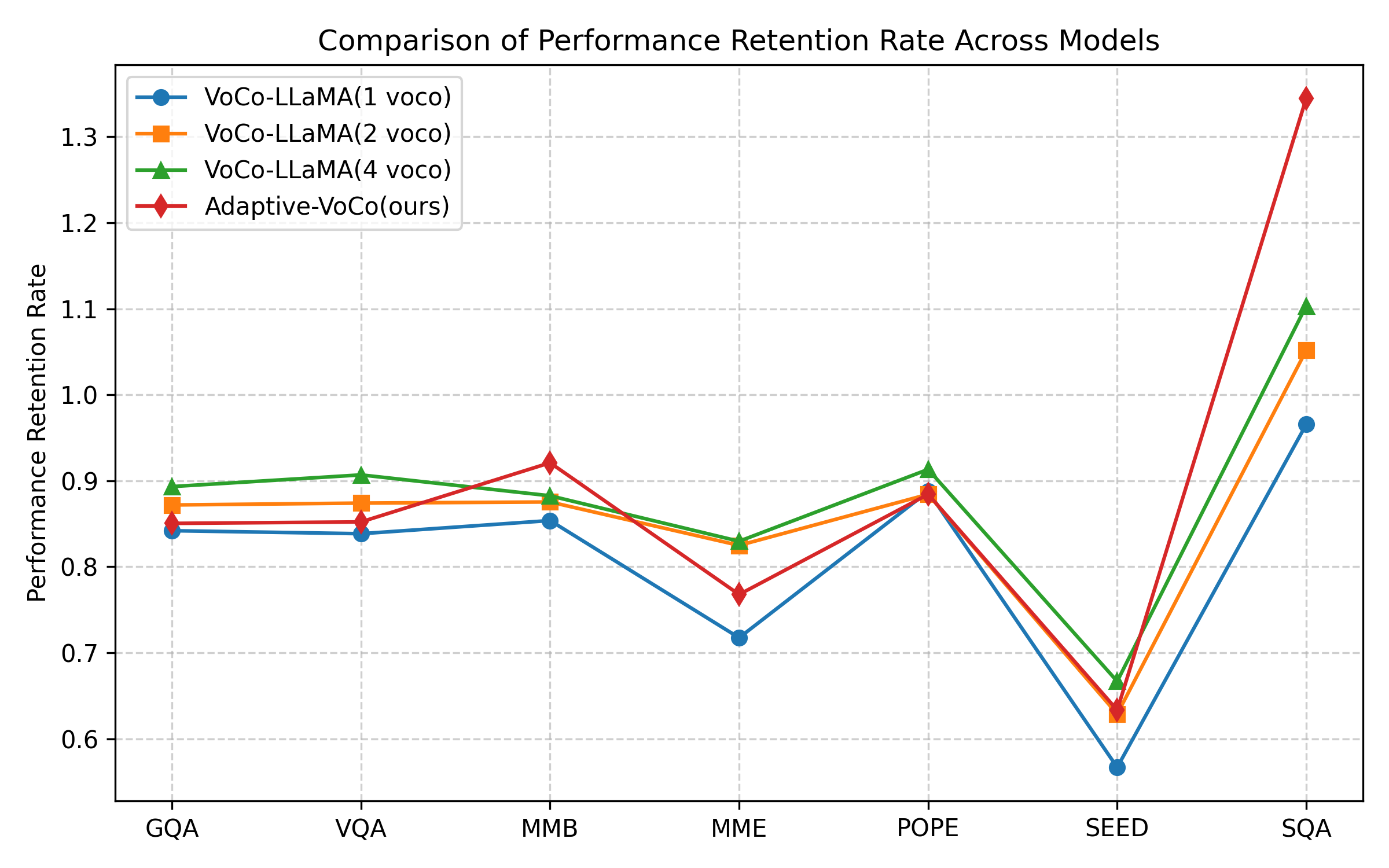}}
  \centerline{(a) Performance comparison}\medskip
\end{minipage}
\hfill
\begin{minipage}[b]{0.49\linewidth}
  \centering
  \centerline{\includegraphics[width=4.0cm]{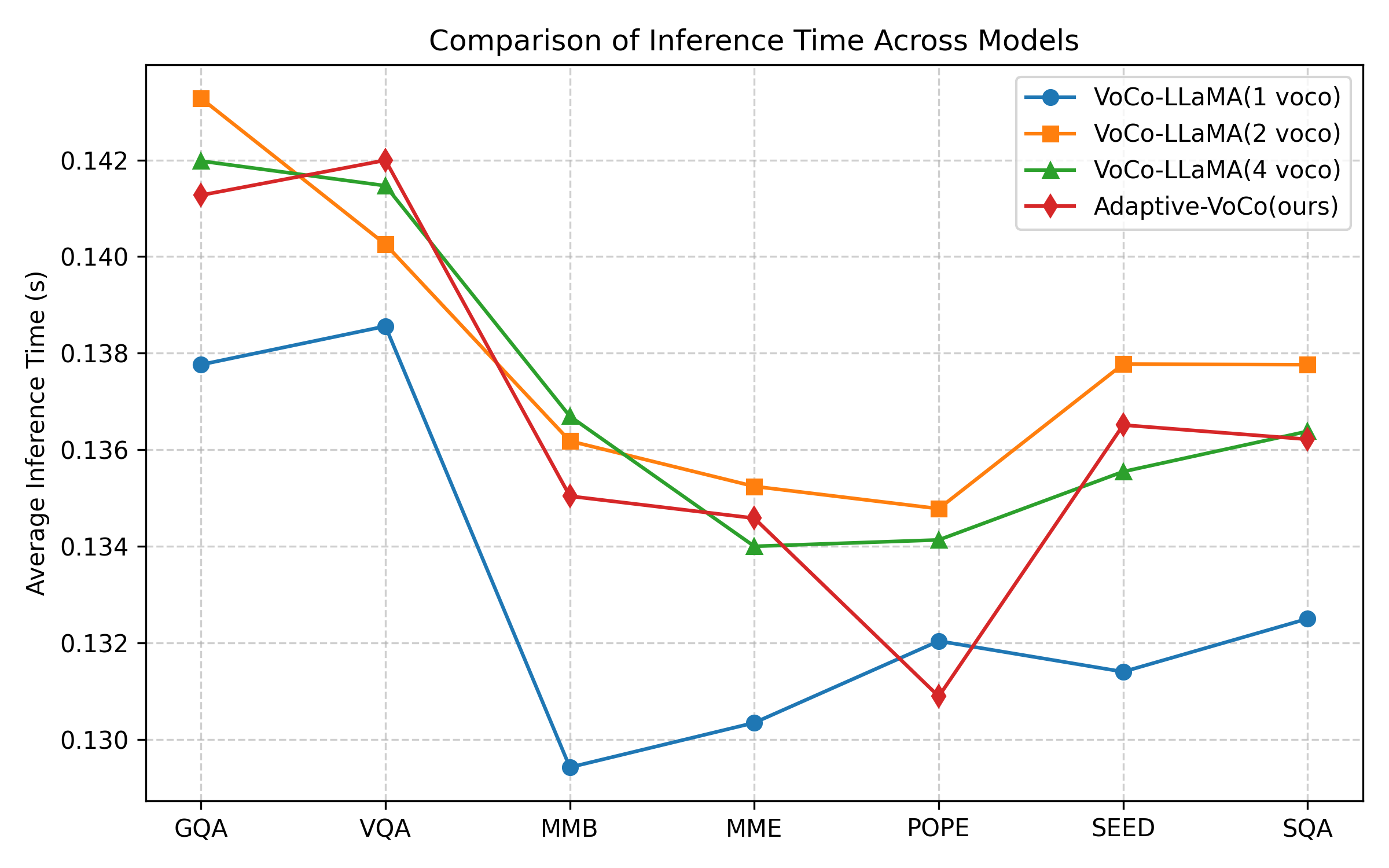}}
  \centerline{(b) Inference time}\medskip
\end{minipage}
\caption{Comparison of Adaptive-VoCo with fixed-rate VoCo-LLaMA models.}
\label{fig:res}
\end{figure}



\subsection{Compression–Performance–Efficiency Trade-off}

To investigate the interplay between visual token compression, downstream performance, and inference efficiency, we compare Adaptive-VoCo with fixed-rate VoCo-LLaMA models across seven multimodal benchmarks.

\textbf{Performance vs. Compression.} Figure~\ref{fig:res}(a) shows the performance retention rates of fixed-rate VoCo-LLaMA models with $K=1,2,4$ tokens and Adaptive-VoCo. As expected, aggressive compression ($K=1$) leads to the lowest retention, while allocating more tokens ($K=4$) significantly improves performance, particularly on complex benchmarks such as MMBench, MME, and SQA$^I$. Our Adaptive-VoCo model consistently achieves comparable or better performance than the $K=2$ and $K=4$ fixed-rate models by dynamically assigning more tokens to challenging inputs and fewer tokens to simpler ones. This demonstrates that complexity-aware token allocation allows the model to maintain high visual understanding across diverse benchmarks without wasting token capacity on easy inputs.

\textbf{Efficiency vs. Compression.} Figure~\ref{fig:res}(b) reports the average single-sample inference time for each benchmark. As expected, models with fewer VoCo tokens ($K=1$) achieve the shortest inference time, while allocating more tokens ($K=2$ or $K=4$) increases computational cost due to the larger number of visual tokens processed by the multimodal transformer. Adaptive-VoCo strikes a balance by dynamically adjusting the number of tokens according to input complexity: simple images receive fewer tokens to maintain fast inference, whereas complex images are assigned more tokens to preserve performance. 
\section{Conclusion}
\label{sec:conclusion}
We presented \textbf{Adaptive-VoCo}, a dynamic token compression framework for Vision-Language Models. With a lightweight Rate Predictor, it allocates tokens by input complexity, preserving fidelity for hard cases while reducing cost on simple ones. On seven benchmarks, Adaptive-VoCo consistently surpasses fixed-rate VoCo-LLaMA in both performance retention and inference efficiency, offering a flexible, resource-aware, and practical solution for modern large-scale models.



\vfill\pagebreak

\bibliographystyle{IEEEbib}
\bibliography{strings,refs}

\begin{thebibliography}{10}

\bibitem{z1}
Jusheng Zhang, Zimeng Huang, Yijia Fan, Ningyuan Liu, Mingyan Li, Zhuojie Yang, Jiawei Yao, Jian Wang, and Keze Wang,
\newblock ``{KABB}: Knowledge-aware bayesian bandits for dynamic expert coordination in multi-agent systems,''
\newblock in {\em Forty-second International Conference on Machine Learning}, 2025.

\bibitem{z2}
Jusheng Zhang, Yijia Fan, Wenjun Lin, Ruiqi Chen, Haoyi Jiang, Wenhao Chai, Jian Wang, and Keze Wang,
\newblock ``{GAM}-agent: Game-theoretic and uncertainty-aware collaboration for complex visual reasoning,''
\newblock in {\em The Thirty-ninth Annual Conference on Neural Information Processing Systems}, 2025.

\bibitem{liu2024llavanext}
Haotian Liu,
\newblock ``Llava-next: Improved reasoning, ocr, and world knowledge,'' January 2024.

\bibitem{Qwen2.5-VL}
Shuai Bai and Keqin Chen,
\newblock ``Qwen2.5-vl technical report,''
\newblock {\em arXiv preprint arXiv:2502.13923}, 2025.

\bibitem{li2023blip}
Junnan Li, Dongxu Li, Silvio Savarese, and Steven Hoi,
\newblock ``Blip-2: Bootstrapping language-image pre-training with frozen image encoders and large language models,''
\newblock in {\em International conference on machine learning}. PMLR, 2023, pp. 19730--19742.

\bibitem{z3}
Jusheng Zhang, Kaitong Cai, Yijia Fan, Jian Wang, and Keze Wang,
\newblock ``Cf-vlm:counterfactual vision-language fine-tuning,'' 2025.

\bibitem{z4}
Jusheng Zhang, Kaitong Cai, Yijia Fan, Ningyuan Liu, and Keze Wang,
\newblock ``{MAT}-agent: Adaptive multi-agent training optimization,''
\newblock in {\em The Thirty-ninth Annual Conference on Neural Information Processing Systems}, 2025.

\bibitem{li2024llama}
Yanwei Li, Chengyao Wang, and Jiaya Jia,
\newblock ``Llama-vid: An image is worth 2 tokens in large language models,''
\newblock in {\em European Conference on Computer Vision}. Springer, 2024, pp. 323--340.

\bibitem{z5}
Jusheng Zhang, Yijia Fan, Zimo Wen, Jian Wang, and Keze Wang,
\newblock ``Tri-{MARF}: A tri-modal multi-agent responsive framework for comprehensive 3d object annotation,''
\newblock in {\em The Thirty-ninth Annual Conference on Neural Information Processing Systems}, 2025.

\bibitem{ye2025voco}
Xubing Ye, Yukang Gan, Xiaoke Huang, Yixiao Ge, and Yansong Tang,
\newblock ``Voco-llama: Towards vision compression with large language models,'' 2025.

\bibitem{z6}
Jusheng Zhang, Kaitong Cai, Xiaoyang Guo, Sidi Liu, Qinhan Lv, Ruiqi Chen, Jing Yang, Yijia Fan, Xiaofei Sun, Jian Wang, Ziliang Chen, Liang Lin, and Keze Wang,
\newblock ``Mm-cot:a benchmark for probing visual chain-of-thought reasoning in multimodal models,'' 2025.

\bibitem{z8}
Jusheng Zhang, Kaitong Cai, Qinglin Zeng, Ningyuan Liu, Stephen Fan, Ziliang Chen, and Keze Wang,
\newblock ``Failure-driven workflow refinement,'' 2025.

\bibitem{jiang2023llmlingua}
Huiqiang Jiang, Qianhui Wu, Chin-Yew Lin, Yuqing Yang, and Lili Qiu,
\newblock ``Llmlingua: Compressing prompts for accelerated inference of large language models,''
\newblock {\em arXiv preprint arXiv:2310.05736}, 2023.

\bibitem{z7}
Jusheng Zhang, Xiaoyang Guo, Kaitong Cai, Qinhan Lv, Yijia Fan, Wenhao Chai, Jian Wang, and Keze Wang,
\newblock ``Hybridtoken-vlm: Hybrid token compression for vision-language models,'' 2025.

\bibitem{zhang2025llava}
Shaolei Zhang,
\newblock ``Llava-mini: Efficient image and video large multimodal models with one vision token,''
\newblock {\em arXiv preprint arXiv:2501.03895}, 2025.

\bibitem{z12}
Jusheng Zhang, Yijia Fan, Kaitong Cai, and Keze Wang,
\newblock ``Kolmogorov-arnold fourier networks,'' 2025.

\bibitem{z10}
Jusheng Zhang, Kaitong Cai, Jing Yang, and Keze Wang,
\newblock ``Learning dynamics of vlm finetuning,'' 2025.

\bibitem{hudson2019gqa}
Drew~A Hudson and Christopher~D Manning,
\newblock ``Gqa: A new dataset for real-world visual reasoning and compositional question answering,''
\newblock in {\em Proceedings of the IEEE/CVF conference on computer vision and pattern recognition}, 2019, pp. 6700--6709.

\bibitem{liu2024mmbench}
Yuan Liu, Haodong Duan, Yuanhan Zhang, et~al.,
\newblock ``Mmbench: Is your multi-modal model an all-around player?,''
\newblock in {\em European Conference on Computer Vision (ECCV)}, 2024.

\bibitem{yin2024survey}
Shukang Yin, Chaoyou Fu, Sirui Zhao, Ke~Li, Xing Sun, Tong Xu, and Enhong Chen,
\newblock ``A survey on multimodal large language models,''
\newblock {\em National Science Review}, vol. 11, no. 12, pp. nwae403, 2024.

\bibitem{li2023evaluating}
Yifan Li, Yifan Du, Kun Zhou, Jinpeng Wang, Wayne~Xin Zhao, and Ji-Rong Wen,
\newblock ``Evaluating object hallucination in large vision-language models,''
\newblock {\em arXiv preprint arXiv:2305.10355}, 2023.

\bibitem{li2023seed}
Bohao Li, Rui Wang, Guangzhi Wang, Yuying Ge, Yixiao Ge, and Ying Shan,
\newblock ``Seed-bench: Benchmarking multimodal llms with generative comprehension,''
\newblock {\em arXiv preprint arXiv:2307.16125}, 2023.

\bibitem{lu2022learn}
Pan Lu and Mishra,
\newblock ``Learn to explain: Multimodal reasoning via thought chains for science question answering,''
\newblock {\em Advances in Neural Information Processing Systems}, vol. 35, pp. 2507--2521, 2022.

\bibitem{goyal2017making}
Yash Goyal and Khot,
\newblock ``Making the v in vqa matter: Elevating the role of image understanding in visual question answering,''
\newblock in {\em Proceedings of the IEEE conference on computer vision and pattern recognition}, 2017, pp. 6904--6913.

\end{thebibliography}

\end{document}